\documentclass[10pt,journal,compsoc]{IEEEtran}
\usepackage[T1]{fontenc}
\usepackage{float}
\usepackage{times}
\usepackage{epsfig}
\usepackage{graphicx}
\usepackage{amsmath}
\usepackage{amssymb}
\usepackage[font=small]{caption}
\usepackage{subcaption}
\usepackage{cite}
\newcommand{\RNum}[1]{\uppercase\expandafter{\romannumeral #1\relax}}

\begin{document}
\title{Fused Deep Neural Networks for Efficient Pedestrian Detection }
\author{Xianzhi Du, Mostafa El-Khamy, Vlad I. Morariu, Jungwon Lee, and Larry Davis \\
{\tt\small xianzhi@umiacs.umd.edu}
}

\maketitle

\begin{abstract}
In this paper, we present an efficient pedestrian detection system, designed by fusion of multiple deep neural network (DNN) systems. Pedestrian candidates are first generated by a single shot convolutional multi-box detector at different locations with various scales and aspect ratios. The candidate generator is designed to provide the majority of ground truth pedestrian annotations  at the cost of a large number of false positives. Then, a classification system using the idea of ensemble learning is deployed to improve the detection accuracy. 
The classification system further classifies the generated candidates based on opinions of multiple deep verification networks and a fusion network which utilizes a novel soft-rejection fusion method to adjust the confidence in the detection results. 
To improve the training of the deep verification networks, a novel soft-label method is devised to assign floating point labels to the generated pedestrian candidates. 
A deep context aggregation semantic segmentation network also provides pixel-level classification of the scene and its results are softly fused with the detection results by the single shot detector. Our pedestrian detector compared favorably to state-of-art methods on all popular pedestrian detection datasets. For example, our fused DNN has better detection accuracy on the Caltech Pedestrian dataset than all previous state of art methods, while also being the fastest. We significantly improved the log-average miss rate on the Caltech pedestrian dataset to $7.67\%$ and achieved the new state-of-the-art. 
%
%
\end{abstract}

\begin{IEEEkeywords} Pedestrian detection, object detection, semantic segmentation, soft-label, soft-rejection, network fusion, ensemble learning, convolutional neural network.
\end{IEEEkeywords}

\section{Introduction}
Object detection is an essential problem in computer vision which aims to detect the locations of semantic objects in videos or digital images. Object detection is widely used in areas such as image retrieval, object identification, video surveillance, etc. Pedestrian detection is a branch of object detection problem which deals with detecting the specific human class. It has applications in various topics such as advanced driver assistance systems, person identification, face recognition, etc. 

The pedestrian detection problem can be decomposed into region proposal generation, feature extraction, and pedestrian verification. In general, object detection involves  generating candidates for bounding boxes enclosing the objects of interest, extracting robust features as high level representations of the candidates, and verifying each candidate to be a true or a false positive. In recent years, convolutional neural network based techniques have successfully been applied to pedestrian detection and achieved better performances in many challenging scenarios. Li et al. \cite{safcnn} trained multiple Fast R-CNN \cite{fastrcnn} based networks to detect pedestrians with different scales and combined results from all networks to generate the final results. Hosang et al. \cite{SCF+AlexNet} used the SquaresChnFtrs \cite{tenyears} method to generate pedestrian proposals and trained AlexNet \cite{alexnet} to perform pedestrian verification. Zhang et al. \cite{rpn} used a Region Proposal Network (RPN) \cite{fasterrcnn} to compute pedestrian candidates and a cascaded Boosted Forest \cite{bf} to perform sample re-weighting to classify the candidates.

\begin{figure*}
\begin{center}
   \includegraphics[width=0.6\linewidth]{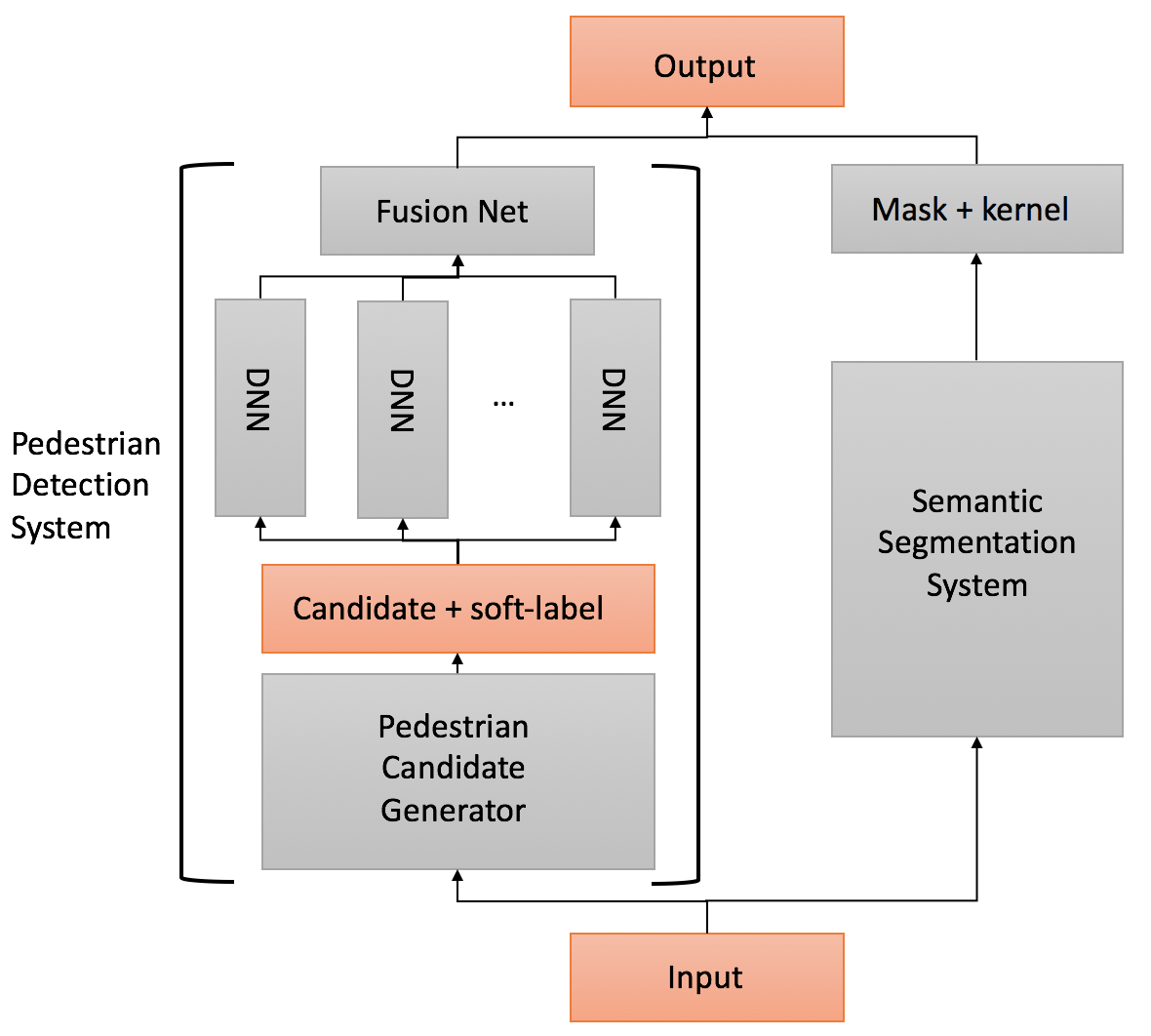}
\end{center}
   \caption{Proposed Fused DNN Architecture}
\label{fig:pipeline}
\end{figure*}

In this paper, we propose a deep neural network fusion architecture to address the pedestrian detection problem, called Fused Deep Neural Network. 
Compared to previous methods, our proposed system is faster while achieving better detection accuracy. The architecture consists of a pedestrian candidate generator, which is obtained by training a deep convolutional neural network trained as a  single shot detector (SSD) with a high detection rate, albeit a large false positive rate. 
A novel soft-rejection strategy is used to adjust the confidence in the detector candidates  by fusion with a classification network employing an ensemble learning approach, and semantic segmentation network which provides pixel-wise labeling. The classification network deploys an ensemble of deep neural networks trained independently as verification networks, and their results are softly fused together with the detection results using the novel soft-rejection method. To prepare the training data for the verification networks, we devise a novel soft-label method to assign floating point labels to the detected candidates. Unlike traditional hard-label method for object verification, where binary labels are used, the value of the our soft-label is set to be the largest overlap ratio between the current detected bounding box and all the ground-truth bounding boxes, and is adjusted to saturate to the binary values.
Additional accuracy improvements can be achieved at the expense of speed by the parallel semantic segmentation network which provides pixel-wise labels to generate a segmentation mask that delivers another soft confidence vote on the generated pedestrian candidates, and are further fused within the soft fusion framework. 
The proposed network architecture is shown in Figure~\ref{fig:pipeline}. 

Some of the ideas in this paper were presented at the 2017 IEEE WACV \cite{fdnn_xianzhidu}. In this paper, we provide more detailed analysis of these ideas, show results on more datasets, and provide additional enhancements that improved the performance over that presented at the 2017 IEEE WACV  \cite{fdnn_xianzhidu}. The new techniques which we present here and helped provide the additional gains over  \cite{fdnn_xianzhidu} are the soft-label method for training classification methods, learning the parameters of soft-rejection fusion by an additional fusion network, and the new kernel based method to fuse the results of the semantic segmentation system and the detection system. The new techniques of this paper helped to significantly increase the detection accuracy on the Caltech dataset from $8.18\%$ to $7.65\%$. We also extend the model to work on more classes besides pedestrians, such as cars and cyclists. Besides the Caltech Pedestrian dataset, we evaluated on three more popular pedestrian detection datasets: INRIA, ETH, and KITTI. Our techniques performed better than all the previous state-of-the-arts on Caltech, INRIA, and ETH in both accuracy and speed, and achieved comparable results on KITTI. More ablation analysis is conducted to explain the effectiveness of our system.

The rest of this paper is organized as follows. Section $2$ provides a detailed description of the pedestrian detection system. Section $3$ describes the semantic segmentation system and how it helps to refine the detection results. Section $4$ discusses the experiment results and explores the effectiveness of each component of the system. Section $5$ draws conclusions on this work.

\section{Pedestrian Detection system}
\subsection{Pedestrian Candidate Generator}
In order to quickly obtain pedestrian candidates in various sizes and aspect ratios at all possible locations of the input image, we use an single shot multi-box detector (SSD) \cite{SSD} as the candidate generator. The main reason we select SSD instead of other systems is that it uses multiple feature maps as the output layers. By lowering the accepting threshold of the confidence score, it outputs a large number of pedestrian candidates which are very likely to cover all the ground-truth pedestrians. Since it has a fully convolutional framework, it's also very fast.

\begin{figure}[thb]
\begin{center}
   \includegraphics[width=0.7\linewidth]{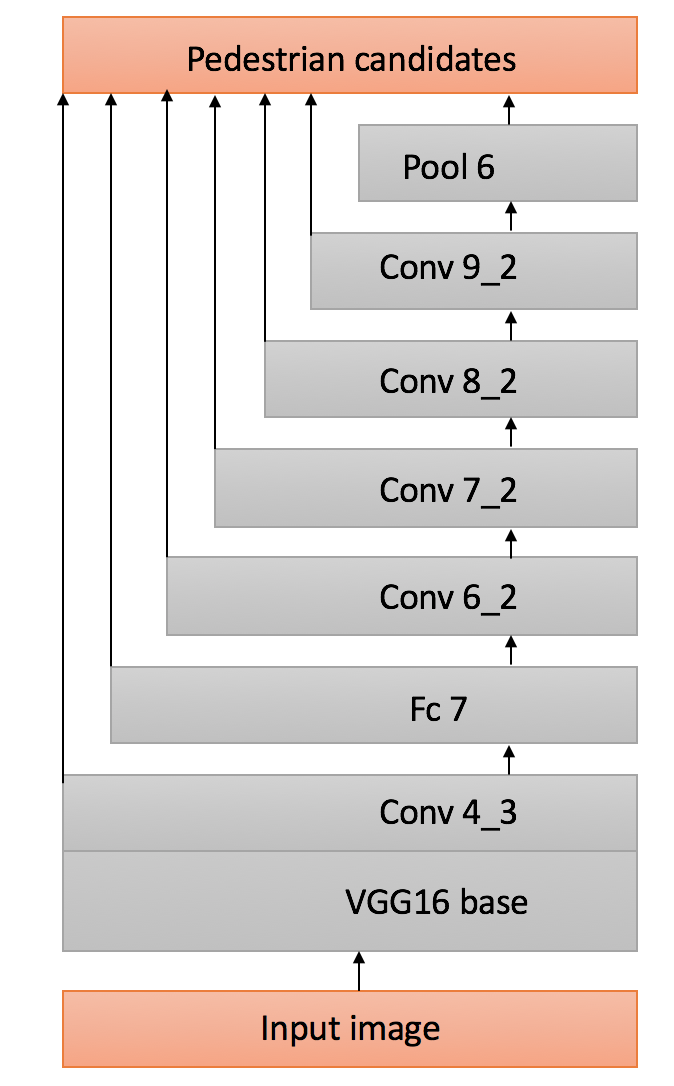}
\end{center}
   \caption{Fully Convolution}
\label{fig:ssd}
\end{figure}

The SSD network is a feed-forward convolutional network which consists of a VGG16 base, 8 convolutional layers above it, and a global pooling layer at the top. In the VGG16 base,  the kernel size of the pool5 layer is set to $3\times 3$ and the stride is set to one, and the fc6 and fc7 layers are converted to dilated convolutional layers. Bounding box regression and classification are performed on the feature maps of 'conv4\textunderscore 3', 'fc7', 'conv6\textunderscore 2', 'conv7\textunderscore 2', 'conv8\textunderscore 2', 'conv9\textunderscore 2', and 'pool6' to generate the pedestrian candidates.
Since the output of 'conv4\textunderscore 3' has a much larger feature scale than that of other filters, an $L_2$ normalization technique is used to scale down the feature magnitudes  \cite{ParseNetLW} . After each output layer, bounding box (BB) regression and classification are performed to generate pedestrian candidates. 
The network architecture of the deployed pedestrian SSD is shown in Figure~\ref{fig:ssd}. 

To generate the pedestrian candidates, a set of default bounding boxes are slided on top of each output feature map.  For each output layer of size $m \times n \times p$, a set of default BBs at different scales and aspect ratios are placed at each location. At every pixel location of the 7 output layers, we place 6 default bounding boxes (BBs) with aspect ratios $[0.1, 0.2, 0.41, 0.8, 1.6, 3.0]$ and relative heights $[0.05, 0.1, 0.24, 0.38, 0.52, 0.66, 0.80]$. Since $0.41$ is the average aspect ratio of all the pedestrian annotations, we place another set of default bounding boxes with that aspect ratio and relative heights $[0.1, 0.24, 0.38, 0.52, 0.66, 0.80, 0.94]$. In the training stage, we further categorize all the pedestrians into three classes: `Full pedestrian', `Occluded pedestrian', and `People'. The `People' class is defined as a group of people that are very close to or overlap with each other. The aspect ratios $0.6$ and $3.0$ are designed for `People'. At each default bounding box, a $3 \times 3 \times p$ convolutional kernel is applied  to produce classification scores, and perform bounding box regression by calculating the location offsets with respect to the default bounding box.

The multi-task training objective of SSD is given by Equation~\eqref{eq:1}
\begin{equation}
L=\frac{1}{N}(L_{conf}+\alpha L_{loc})
\label{eq:1}
\end{equation}
where $L_{conf}$ is the softmax loss for the classification task and $L_{loc}$ is the smooth $L_1$ localization loss \cite{fastrcnn}, $N$ is the number of positive default boxes, and $\alpha$ is a constant weighting factor to keep a balance between the two losses.  Since SSD uses 7 output layers to generate multi-scale BB outputs, it provides a large pool of pedestrian candidates varying in scales and aspect ratios. 

When training the SSD as the primary candidate generator, a default detection BB is labeled as positive if it has a Jaccard overlap (intersection over union ratio) greater than 0.5 with any ground truth BB, otherwise it is labeled as negative. 
The SSD generated pedestrian candidates should cover almost all ground truth pedestrians, while being as accurate as possible. This is essential to our fused DNN architecture, where following classification networks will attempt to improve the precision by decreasing the confidence in many of the false positives introduced by the SSD, while preserving the recall rate. By being a fully convolutional network, SSD is a fast candidate generator.

\subsection{Classification System \label{FNN}}
\subsubsection{Classification networks with soft-label method}
In this section, we describe the soft-label method we devised for preparing the training data for the classification network. The hard-label used in common object detection networks assigns a binary label to each pedestrian candidate bounding box by thresholding the Jaccard overlap ratio between this bounding box and the ground-truth bounding boxes. However, this is not the optimal strategy, especially when the overlap ratio is close to the threshold. In this work, we introduce the soft-label method to label pedestrian candidates for further classification. The soft-label method will assign a floating point label to each pedestrian candidate using the largest overlap ratio between the current pedestrian candidate and all the ground-truth bounding boxes. Suppose we have a pedestrian candidate and the ground-truth bounding box it overlaps with the most. The soft-labels for the pedestrian class $\mbox{label}_{ped}$ and the background class $\mbox{label}_{bg}$ are respectively calculated by Equations~\eqref{eq:2} and~\eqref{eq:3},  where $A_{BB_d}$ and $A_{BB_g}$ represent the areas covered by the detection BB and the ground truth BB, respectively

\begin{equation} \label{}
    \mbox{label}_{ped}= \frac{A_{BB_{d}}\cap A_{BB_{g}}}{A_{BB_{d}}}
\label{eq:2}
\end{equation}
\begin{equation} \label{}
    \mbox{label}_{bg}=1- \frac{A_{BB_{d}}\cap A_{BB_{g}}}{A_{BB_{d}}}
\label{eq:3}
\end{equation}

However, we found that it is better to use a soft-label in cases of mild confidence, and use a hard-label in cases when the confidence in the label exceeds a threshold. Hence, we devised a hybrid soft-label method as follows: If the overlap ratio between the pedestrian candidate and the ground truth bounding box is lower than a threshold $th_a$ or greater than a threshold $th_b$, we think it's safe to label the current sample as background or pedestrian with probability one. For the candidates with intermediate confidence, the soft-label method is used to assign floating point labels, and we normalize the range [$th_a$, $th_b$] to [$0$, $1$]. Moreover, for convenience of implementation, the soft-label is made continuous over its range from $0$ to $1$. The final soft-label method is shown by Equations~\eqref{eq:6} and~\eqref{eq:7}. 

\begin{equation}
    \mbox{label}_{ped}=
\begin{cases}
    1,& \text{if } \frac{A_{BB_{d}}\cap A_{BB_{g}}}{A_{BB_{d}}}>th_b \\         
    0,& \text{if } \frac{A_{BB_{d}}\cap A_{BB_{g}}}{A_{BB_{d}}}<th_a \\ 
    \frac{\frac{A_{BB_{d}}\cap A_{BB_{g}}}{A_{BB_{d}}}-th_a}{th_b-th_a}, & \text{otherwise,}
\end{cases}
\label{eq:6}
\end{equation}
\begin{equation}
\mbox{label}_{bg}=1-\mbox{label}_{ped}.
\label{eq:7}
\end{equation}

The classification networks are trained to minimize the cross entropy loss objective function, 
\begin{equation} \label{}
\epsilon=-\sum^{c}_{i=1}l_i \log{y_i}
\label{eq:4}
\end{equation}
where $c$ is the number of classes, $l_i$ and $y_i$ are the soft-label and the softmax probability for class $i$. Note that $\sum_i l_i=1$.
Note that for the conventional binary labeling method, $l_i$ is the indicator function, which is $1$ for the correct class and $0$ otherwise. Minimizing the cross entropy is equivalent to maximizing the softmax probability of the correct class. In our soft-label case, the softmax probabilities of all the classes are used to contribute to the cross entropy loss. The floating point soft-labels will determine how much each class contributes. 
When doing back-propagation, the derivative of the cross entropy cost function with respect to class $i$ is calculated as Equation~\eqref{eq:5}.

\begin{equation} \label{}
\begin{aligned}
\frac{\partial \epsilon}{\partial z_i}&=-\sum^{c}_{j=1}l_j\frac{\partial \log{y_j}}{\partial z_i}=-\frac{l_i}{y_i}\frac{\partial y_i}{\partial z_i}-\sum^c_{j\neq i}\frac{l_j}{y_j}\frac{\partial y_j}{\partial z_i}\\
&=-\frac{l_i}{y_i}y_i(1-y_i)-\sum^c_{j\neq i}\frac{l_j}{y_j}(-y_j y_i)\\
&=-l_i+y_i\sum^c_{j=1}l_j=y_i-l_i
\end{aligned}
\label{eq:5}
\end{equation}
We note that Equation~\eqref{eq:5} shows the gradient calculations and holds for both conventional hard-label method as well as for the proposed soft-label method. For conventional hard-label method, a training sample has label $1$ for the correct class, and label $0$ for the incorrect classes. We note that it holds for the soft-label method as long as the sum of the soft-labels over all classes is $1$. 

\begin{figure*}
\begin{subfigure}{.5\textwidth}
  \centering
  \includegraphics[width=.55\linewidth]{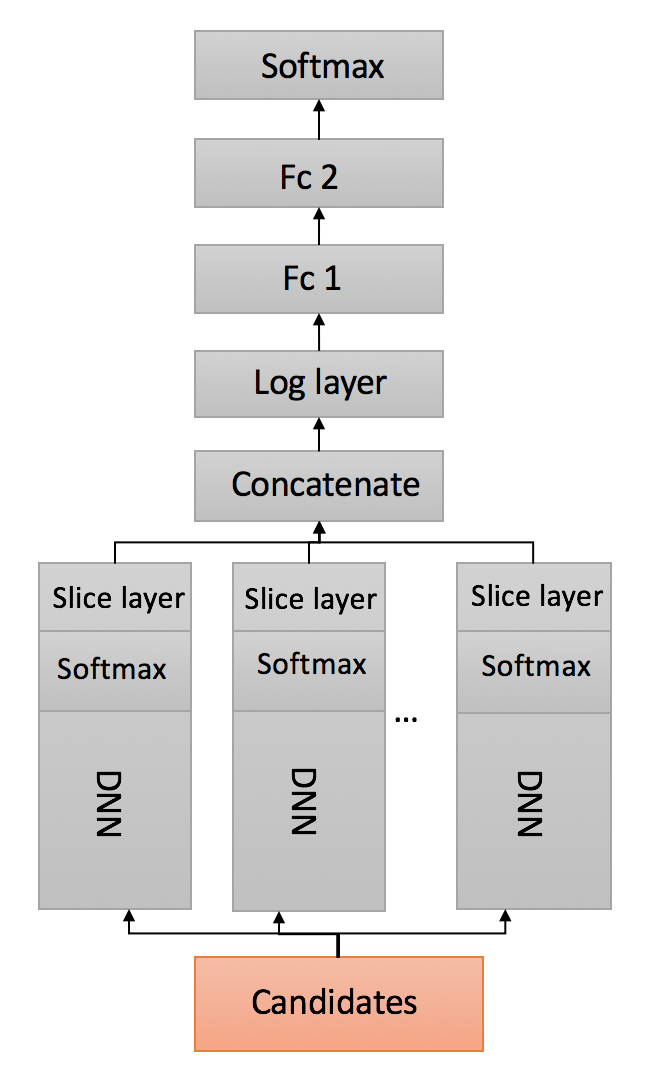}
  \label{fig:sfig1}
\end{subfigure}
\begin{subfigure}{.5\textwidth}
  \centering
  \includegraphics[width=.55\linewidth]{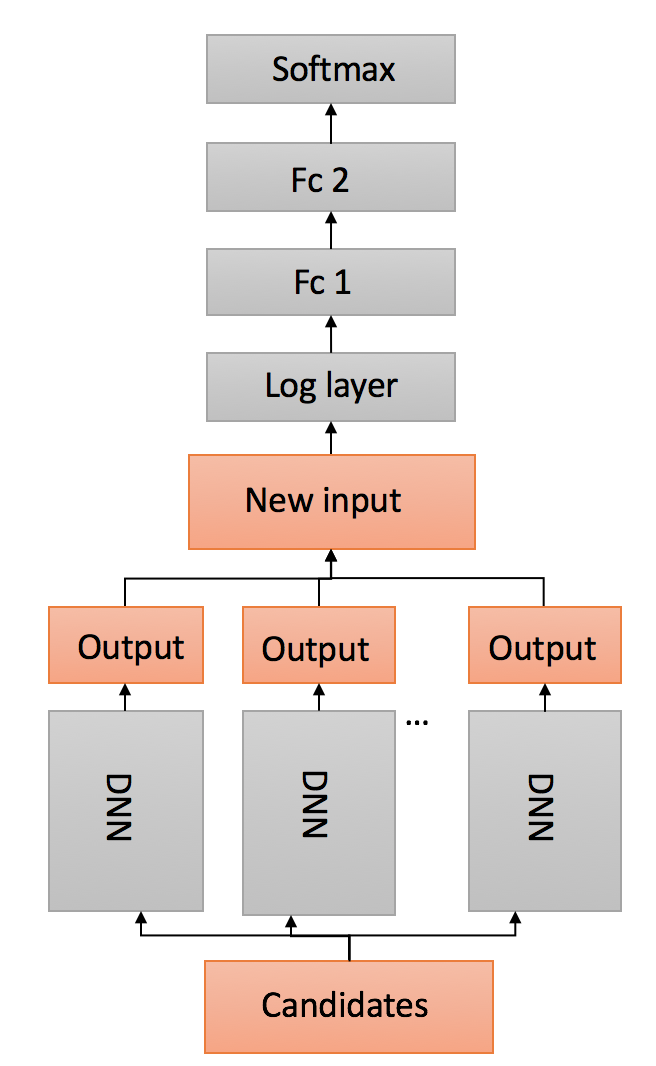}
  \label{fig:sfig2}
\end{subfigure}
   \caption{The two fusion network designs described in the paper. The left structure is an end-to-end training scheme. The right structure trains all the networks separately.}
\label{fig:fusion}
\end{figure*}

As shown in Figure~\ref{fig:fusion}, the classification network constitutes of multiple networks, run in parallel, where we used the idea of ensemble learning. The constituent networks are  deep classification neural networks which have different network structures, but trained with the same input data.
All pedestrian candidates generated by the primary candidate generator with confidence score greater than 0.01 and height greater than 40 pixels are collected as the new training data for the classification network. To solve the problem of classifying candidates with different aspect ratios and sizes, all candidate BBs are rescaled to a fixed input size. The goal of these secondary classification networks is to further classify the detections from the first stage single shot detector as a true detection or a false detection.

\subsubsection{Soft-rejection based fusion \label{subsection:SNF1} }

 The opinions of all the constituent classification networks are fused with those of the candidate generator (CG). By doing so, it's more likely to get a lower error than having a single network. Since it's hard to bias towards each of the single network, it's also less likely to over-fit. Also, since the networks are run in parallel, the speed of the classification network is limited by its slowest constituent network.
There are several conventional methods commonly used for opinion fusion, such as computing the mean of all results, majority voting, or the hard-rejection method. The hard-rejection method will eliminate a pedestrian candidate based on a single negative vote from one classification network. Instead, we introduce the soft-rejection network fusion (SNF) method, as we use classification networks with different structures, and we expect each network to work well in some of the subcategories while performing mediocre in other categories. The soft-rejection based fusion method can be described as follows: For one pedestrian candidate, the $k_{th}$ classification network gives us a classification probability $p_k$. If $p_k$ is higher than a threshold $t_1$, we generate a scaling factor $s_k$ greater than one to boost the initial confidence score generated by the SSD candidate generator. Otherwise, we generate a scaling factor less than one to decrease the initial confidence score. To prevent any of the classification networks from dominating the final results, we set a lower bound $t_2$ to the scaling factors. The scaling factors coming out from all the classification networks are further multiplied together with the initial confidence score to produce the final score. The idea behind this is that instead of accepting or rejecting any candidate, we boost or decrease their scores instead. This is because a poor classification network can be compensated by other good ones with SNF, whereas a wrong elimination of a true pedestrian by hard-rejection cannot be corrected. The SNF idea is illustrated in Equation~\eqref{eq:8} and Equation~\eqref{eq:9}.

\begin{equation}
s_{k}=\max \left(p_{k}\times \frac{1}{t_1},t_2 \right)
\label{eq:8}
\end{equation}
\begin{equation}
S_{FDNN}=S_{CG}\times \prod_{k=1}^{K}s_k. 
\label{eq:9}
\end{equation}

\subsubsection{Soft-rejection based fusion network \label{subsection:SNF2} }
The values of all the parameters in the soft-rejection based fusion method as described in Subsection \ref{subsection:SNF1}  were selected by cross-validation in our previous work \cite{fdnn_xianzhidu}. However, we found that the parameters when optimized on one dataset do not easily generalize to other datasets. Instead of hand-crafting such parameters, we propose in this paper to use a neural network to learn the optimal parameters,  while keeping the idea of the soft-rejection based fusion method. We call this new method soft-rejection based fusion network. 

Let $p_1, p_2, ..., p_K$ be the inputs to the fusion network, where $p_k$ is the softmax output of the $k_{th}$ classification network. The input layer also deploys a log layer to get the  classification log probabilities of the individual networks. The input layer is followed by two fully connected layers, each has $500$ neurons, and one softmax layer to predict the weights for each classification network.  The output of the neural network is the exponent of the weighted sum of all classification log probabilities. This results in a soft-fusion scheme which scales the candidate generator's confidence scores with the  product of (exponentially) weighted classification probabilities of all individual classification networks, where the weights $w_k$ are optimized by the fusion network, and adheres to our soft network fusion framework described by Equation~\eqref{eq:9}, 
\begin{equation}
\begin{aligned}
S_{FDNN2}&=S_{CG}*\exp\left(\sum_{k=1}^K w_k*\log(p_k)\right) \\
&=S_{CG}*\prod_{k=1}^K p_k^{w_k}.
\end{aligned}
\label{eq:10}
\end{equation}

\subsubsection{Training the classification system}
There are two ways to train the classification system. The first method is to train an end-to-end system. For all the classification networks, we remove their loss layers and concatenate the output neurons for the pedestrian class from the softmax layers to form the input layer to the fusion network. This system has classification networks as branches and merged together by the fusion network at the end. This is shown at the left of Figure~\ref{fig:fusion}. However, since we train all the networks together, the structure grows huge and is prone to overfitting. What's more, since all the branches have different structures, they require different settings of optimal hyper-parameters and have different converging speed.

The second method is to train the classification networks first, and use the output probabilities as inputs to train the fusion network separately. Since this is straightforward and easy to implement, as shown at the right figure of Figure~\ref{fig:fusion}, this method is finally used in this paper. 

\section{Pixel-wise Semantic Segmentation System } 

Based on the idea of fusion of multiple experts, we propose to enhance the accuracy of the system by adding an expert network which is trained independently from the candidate generator. In this implementation, we chose the independent network to be a semantic segmentation network trained to provide pixel-wise classification of pixels, in contrast to the region classification done by the candidate generator. 
 In our work, we utilize a semantic segmentation (SS) network based on deep dilated convolutions and context aggregation \cite{sspaper} running in parallel with the pedestrian detection system to further refine the end detection results of the whole system. The SS network is trained on the Cityscapes dataset for driving scene segmentation \cite{cityscapes}. To perform dense prediction, the network consists of a fully convolutional VGG16 network, adapted with dilated convolutions as the front end prediction module, whose output is fed to a multi-scale context aggregation module, consisting of a fully convolutional network whose convolutional layers have increasing dilation factors. We consider both the "person" and "rider" classes in Cityscapes dataset as pedestrians, and the remaining classes as background. 
\subsection{Soft Fusion with a Pixel-wise Semantic Segmentation System \label{subsection:FSS}} 
The results of the semantic segmentation system are fused with those of the pedestrian detection system as follows: First, we process the same original input image using the semantic segmentation system. This produces a binary mask where the foreground pixels are set to $1$ to represent the class of interest (e.g. pedestrian) and the background pixels are set to $0$. Then, for each of the bounding boxes detected produced by the candidate generator, we analysis its pixels at the same locations on the binary mask. The soft fusion scaling factor is computed as the weighted sum of the foreground pixels within the bounding box, where the weighting factors are calculated from a weight matrix, called the Kernel, within the bounding box. To enable this weighted sum, all detected bounding boxes and their overlapped masks are rescaled to have the same size as the Kernel. The Kernel is trained as the mean of the semantic segmentation binary masks of all ground-truth pedestrian bounding boxes in the training set and normalized to have a sum  of $1$. This fusion method is illustrated in Equations~\eqref{eq:11} and~\eqref{eq:12}.

\begin{figure*}
\begin{center}
   \includegraphics[width=1.0\linewidth]{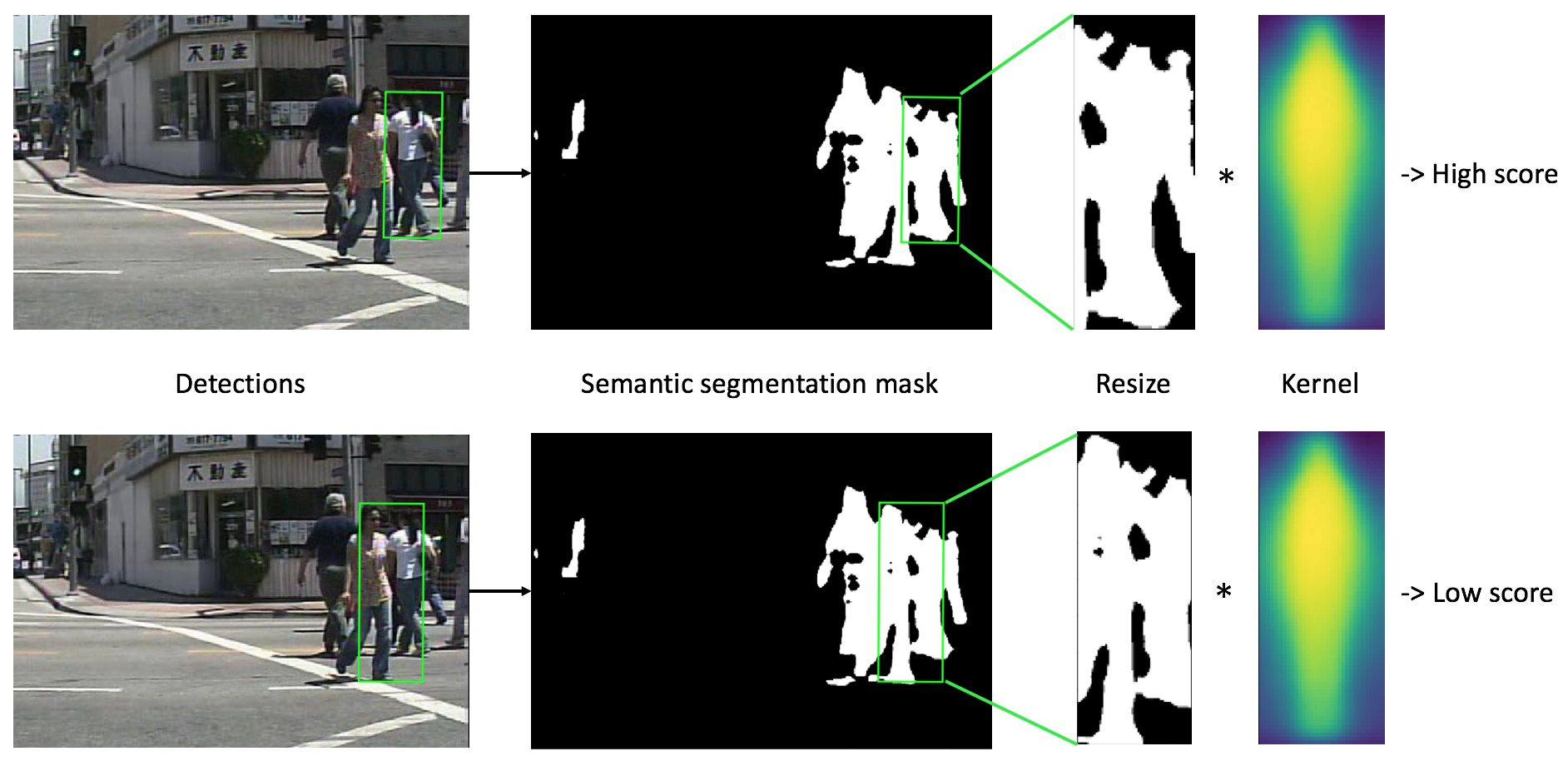}
\end{center}
   \caption{The idea of kernel-based method to fuse the semantic segmentation system and the detection system.}
\label{fig:kernel}
\end{figure*}

\begin{equation}
S_{SS}=\sum_{i,j}^{A_{BB}} \mbox{mask}(i,j)  \times \mbox{Kernel}(i,j),
\label{eq:11}
\end{equation}
\begin{equation}
S_{FDNN2 + SS}=S_{FDNN2} \times S_{SS}
\label{eq:12}
\end{equation}
where the $A_{BB}$ is the area of the bounding box. $\mbox{mask}(i,j)$ and $\mbox{Kernel}(i,j)$ are the pixel value of the binary mask and the Kernel at location $(i, j)$. From the visualization of the weight mask, illustrated in Figure~\ref{fig:kernel}, we see that the pixels at the center of the kernel tend to have higher values than the pixels at the boundary. This agrees with the fact that in a perfect detection, the object of interest tends to appear at the center of the bounding box. We can see that the Kernel will have the effect of boosting the score of a detection whose bounding box fits the object of interest (e.g. pedestrian) and decreasing the score of a detection whose bounding box is not well located. This model will be referred to as `FDNN2 + SS'

It is worth noting that in our previous work \cite{fdnn_xianzhidu}, the soft fusion of the semantic segmentation results with that of the object detection system, labeled `FDNN + SS', was done differently. 
For the sake of completeness and comparison of the results, we describe it here. 
 The SS mask is intersected with all detected BBs produced by the CG and the degree to which each candidate's BB overlaps with the pedestrian category in the SS activation mask gives a measure of the confidence of the SS network in the candidate generator's results. If the pedestrian pixels occupy at least $20\%$ of the candidate BB area, we accept the candidate and keep its score unaltered; Otherwise, the candidate generator's  scores are softly fused by scaling as in Equation~\ref{eq:FusedDNNSS},
\begin{equation} \label{eq:FusedDNNSS}
    S_{FDNN+SS}=
\begin{cases}
    S_{FDNN},& \text{if } \frac{A_M}{A_{BB}}>0.2 \\ 
    S_{FDNN}\times \max(\frac{A_M}{A_{BB}} \times a_{ss}, & b_{ss}),  \text{otherwise}
\end{cases}
\end{equation}
where $A_{BB}$ represents the area of the BB, $A_M$ represents the area within $A_{BB}$ covered by the semantic segmentation mask, $a_{ss}$, and $b_{ss}$ are chosen as $4$ and $0.35$ by cross validation. 

We also note that SNF of the CG network with an SS network is slightly different from SNF with a classification network. The reason is that the SS network can generate new detections which have not been produced by the CG, which is not the case for the classification networks. To address this, the proposed SNF methods `FDNN2 + SS' and `FDNN + SS' eliminate new detections from the SS network, which do not overlap with any CG detection.

\section{Experiments and result analysis}
\subsection{Training settings}
The proposed method is trained on the training sets of the Caltech Pedestrian dataset, the ETH dataset, and the TudBrussels dataset.

To train the pedestrian candidate generator, both the original images and the horizontally flipped images which contain at least one annotated bounding box are used, which results in around $68,000$ training images in total. Among all the annotated bounding boxes, there are about $109,000$ annotated bounding boxes in 'Person\textunderscore full' class, $60,000$ annotated bounding boxes in 'Person\textunderscore occluded' class, and $35,000$ bounding boxes in 'People' class. All the images are of size  $480\times 640$. The model is fine-tuned from the Microsoft COCO \cite{coco} pre-trained SSD model for $40,000$ iterations using the standard stochastic gradient descent (SGD) algorithm and the back-propagation algorithm at a learning rate of $10^{-5}$.

To train the classification system, all the ground-truth annotations and the pedestrian candidates generated from the previous stage with height greater than $40$ pixels and confidence score larger than $0.01$ are selected, and rescaled into a fixed size of $250\times 250$ to represent the training samples. For data augmentation,  a $224\times 224$ patch is randomly cropped out of each training sample and horizontally flipped with probability $0.5$. To label the training samples, the soft-label method as described by Equations~\eqref{eq:6} and~\eqref{eq:7}
 is implemented. The thresholds $th_a$ and $th_b$ are set to $0.4$ and $0.6$, respectively. To build the classification networks, one ResNet-50 \cite{res50} and one GoogleNet \cite{googlenet} are used as the classification networks. Both of the classifiers are fine-tuned from the ImageNet pre-trained models using the standard SGD algorithm and the back-propagation algorithm at a learning rate of $10^{-4}$.

As we don't have pixel-level labels for pedestrian detection datasets
to incorporate the semantic segmentation network, the dilated convolution model \cite{sspaper} trained on the Cityscapes dataset \cite{cityscapes} is directly implemented. All the classes are considered as background, except the 'Person' and 'Rider' classes which are considered as the `Pedestrian' class. Due to the lack of well-labeled pedestrian datasets for our problem, no fine-tuning is involved in this step. All the input images are rescaled from $480\times 640$ into $1024\times 2048$. To preserve the aspect ratio so as to preserve the human body shape, the image's height is firstly scaled to $1024$ and then blank patches are padded on both left and right sides. 

All the above mentioned models are built with the Caffe deep learning framework \cite{caffe}.

\subsection{Evaluation settings and results}
We evaluate the proposed method on the four most popular pedestrian detection datasets: the Caltech Pedestrian dataset, the INRIA dataset, the ETH dataset, and the KITTI dataset. The log-average miss rate (L-AMR) is used as the performance evaluation metric \cite{caltech}. L-AMR is computed evenly spaced in log-space in the range $10^{-2}$ to $10^0$ by averaging the miss rate at the rate of nine false positives per image (FPPI) \cite{caltech}. There are multiple evaluation settings defined based on the height and visible part of the bounding boxes. The most popular settings are listed in Table~\ref{tab:cal_settings}. 
%

\begin{table}[h!]
\begin{center}
\begin{tabular}{|l|l|l|}
\hline
Setting & Description\\
\hline\hline
Reasonable & 50+ pixels. Occ. none or partial\\
All & 20+ pixels. Occ. none, partial, or heavy\\
Far & 30- pixels\\
Medium & 30-80 pixels\\
Near & 80+ pixels\\
Occ. none & 0\% occluded\\
Occ. partial & 1-35\% occluded\\
Occ. heavy & 35-80\% occluded\\
\hline
\end{tabular}
\end{center}
\caption{Evaluation settings for Caltech Pedestrian dataset.}
\label{tab:cal_settings}
\end{table}

We refer to the new models of this paper as F-DNN2, which is the proposed pedestrian detection system with a fusion network, and F-DNN2+SS, which is F-DNN2 system fused with the semantic segmentation system, and to the models of our previous work \cite{fdnn_xianzhidu} as FDNN and FDNN+SS, for fusion of the CG with the classification system only or with both the classification system and the SS network, respectively, as described in Subsections~\ref{subsection:SNF1},~\ref{subsection:SNF2}, and~\ref{subsection:FSS}. Descriptions of each dataset and its evaluation results are given below.

\begin{table*}[ht]
\begin{center}
\begin{tabular}{|l|l|l|l|l|l|l|l|l|}
\hline
Method & Reasonable & All & Far & Medium & Near & Occ. none & Occ. partial & Occ. heavy\\
\hline\hline
SCF+AlexNet \cite{SCF+AlexNet} & 23.32\% & 70.33\% & 100\% & 62.34\% &  10.16\% & 19.99\% & 48.47\% & 74.65\%\\
SAF R-CNN \cite{safcnn} & 9.68\% & 62.6\% & 100\% & 51.8\% & \textbf{0\%} & 7.7\% & 24.8\% & 64.3\%\\
MS-CNN \cite{mscnn} & 9.95\% & 60.95\% & 97.23\% &  49.13\% & 2.60\% & 8.15\% & 19.24\% & 59.94\%\\
DeepParts \cite{DeepParts2015} & 11.89\% & 64.78\% & 100\% & 56.42\% & 4.78\% & 10.64\% & 19.93\% & 60.42\%\\
CompACT-Deep \cite{CompACT2015} & 11.75\% & 64.44\% & 100\% & 53.23\% & 3.99\% & 9.63\% & 25.14\% & 65.78\%\\
RPN+BF \cite{rpn} & 9.58\% & 64.66\% & 100\% & 53.93\% & 2.26\% & 7.68\% & 24.23\% & 69.91\%\\
F-DNN+SS \cite{fdnn_xianzhidu} & 8.18\% & 50.29\% & 77.47\% & \textbf{33.15\%} & 2.82\% & 6.74\% & \textbf{15.11\%} & 53.76\% \\
F-DNN2 (ours) & 8.12\% & 51.86\% & 77.99\% & 36.72\% & 1.68\% & 6.75\% & 17.51\% & 40.84\% \\
F-DNN2+SS (ours) & \textbf{7.67\%} & \textbf{49.80\%} & \textbf{75.83\%} & 35.09\% & 1.51\% & \textbf{6.35\%} & 16.17\% & \textbf{39.84\%} \\
\hline
\end{tabular}
\end{center}
\caption{Detailed breakdown performance comparisons of our models and other state-of-the-art models on the 8 evaluation settings. All numbers are reported in L-AMR.}
\label{tab:cal}
\end{table*}
\begin{table*}[ht!]
\begin{center}
\begin{tabular}{|l|l|l|l|l|l|l|l|}
\hline
Method & RPN+BF & SketchTokens & SpatialPooling & RandForest & VJ & HOG & F-DNN2+SS (ours)  \\
\hline\hline
L-AMR & 6.88\% & 13.32\% & 11.22\% &  15.37\% & 72.48\% & 63.49\%& \textbf{6.78}\%\\
\hline
\end{tabular}
\end{center}
\caption{Performance comparisons of our models and other state-of-the-art models on the INRIA dataset.}
\label{tab:inria}
\end{table*}
\begin{table*}[ht]
\begin{center}
\begin{tabular}{|l|l|l|l|l|l|l|l|}
\hline
Method & RPN+BF & TA-CNN & SpatialPooling & RandForest & VJ & HOG & F-DNN2+SS (ours)  \\
\hline\hline
L-AMR & 30.32\% & 34.98\% & 37.37\% &  45.04\% & 74.69\% & 89.89\%& \textbf{30.02}\%\\
\hline
\end{tabular}
\end{center}
\caption{Performance comparisons of our models and other state-of-the-art models on the ETH dataset.}
\label{tab:eth}
\end{table*}
\begin{table*}[ht]
\begin{center}
\begin{tabular}{|l|l|}
\hline
Setting & Description\\
\hline\hline
Easy & Min. BB height: 40 Px, Max. occlusion level: Fully visible, Max. truncation: 15\%\\
Moderate & Min. BB height: 25 Px, Max. occlusion level: Partly occluded, Max. truncation: 30\%\\
Hard & Min. BB height: 25 Px, Max. occlusion level: Difficult to see, Max. truncation: 50\%\\
\hline
\end{tabular}
\end{center}
\caption{Evaluation settings for KITTI object detection dataset.}
\label{tab:kitti_settings}
\end{table*}

\begin{figure}
\begin{center}
   \includegraphics[width=1\linewidth]{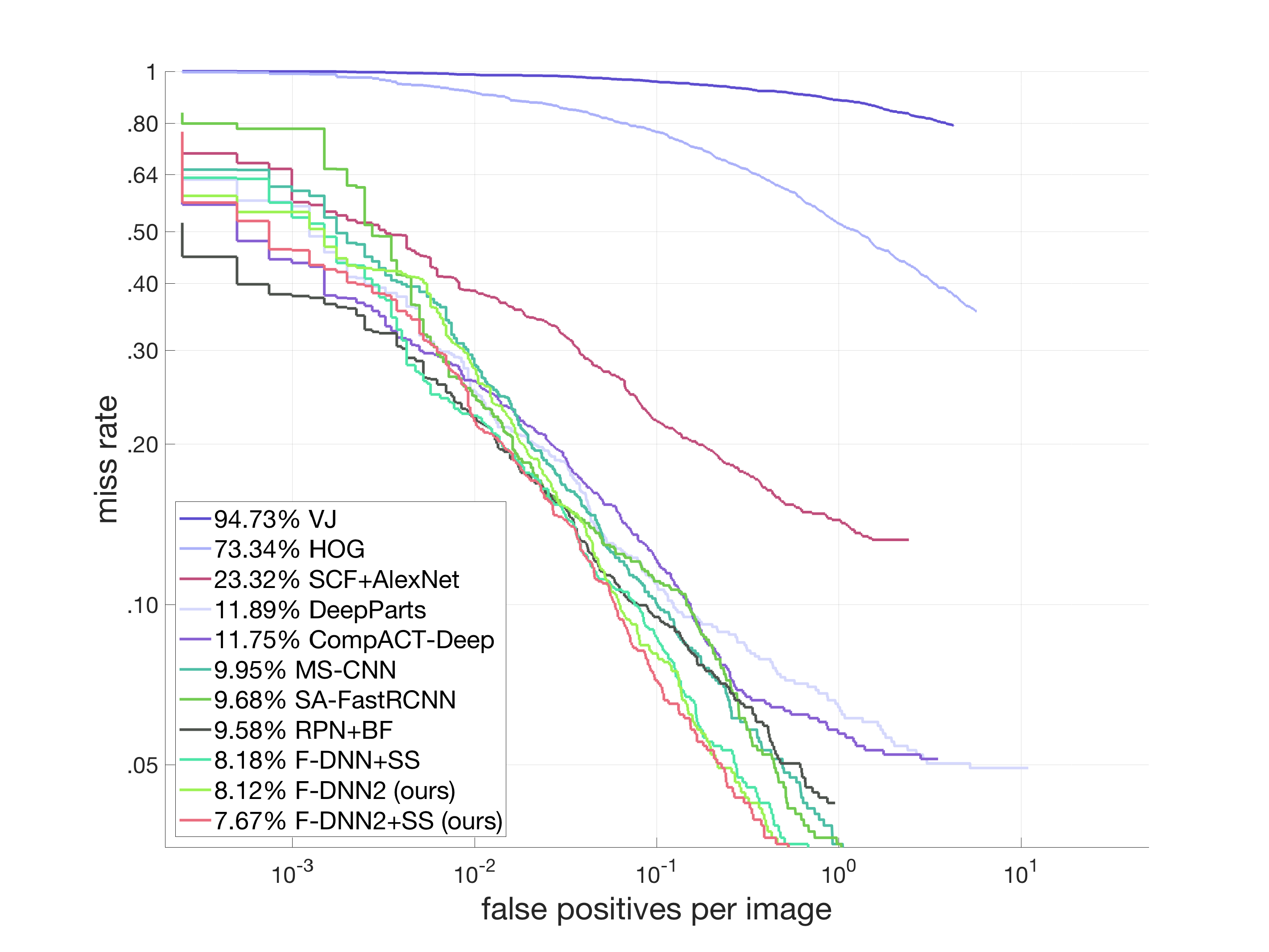}
\end{center}
   \caption{L-AMR vs. FPPI plot under the 'Reasonable' evaluation setting on Caltech Pedestrian dataset.}
\label{fig:cal_res}
\end{figure}

\begin{figure}
\begin{center}
   \includegraphics[width=1\linewidth]{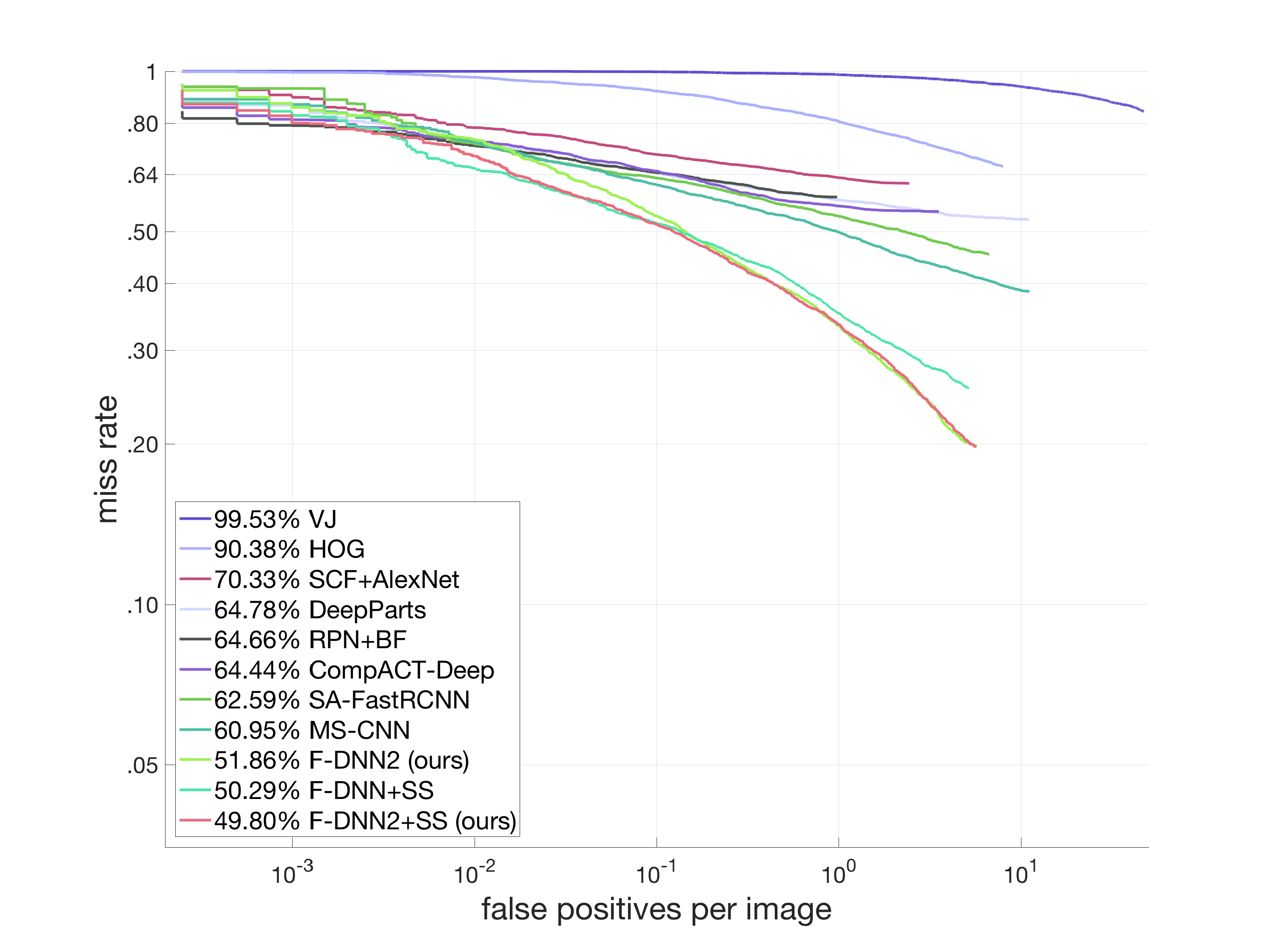}
\end{center}
   \caption{L-AMR vs. FPPI plot under the 'All' evaluation setting Caltech Pedestrian dataset.}
\label{fig:cal_all}
\end{figure}

\textbf{Evaluation on the Caltech Pedestrian data}: The Caltech Pedestrian dataset contains 11 sets (S0-S10), where each set consists of 6 to 13 one-minute long videos collected from a vehicle driving through an urban environment. There are about $250,000$ frames with about $350,000$ annotated BBs and $2,300$ unique pedestrians. Each bounding box is assigned with one of the three labels: 'Person', 'People' (large group of individuals), and 'Person?' (unclear identifications). The detailed breakdown performances of our two models (detection system only and detection system plus semantic segmentation system) on this dataset is shown in Table~\ref{tab:cal}. We compare with all the state-of-the-art methods reported on Caltech Pedestrian website. We can see that both of our models significantly outperform others on almost all evaluation settings. On the 'Reasonable' setting, our `FDNN+SS' best model achieves $8.18\%$ L-AMR, which has a $14.6\%$ relative improvement from the previous best result $9.58\%$ by RPN+BF. Even more accuracy can be obtained by our proposed `FDNN2' and `FDNN2+SS' models, which achieve $8.12\%$ and $7.67\%$ L-AMR, respectively. On the 'All' evaluation setting, we achieve $50.29\%$ with `FDNN+SS', a relative improvement of $17.5\%$ from $60.95\%$ by MS-CNN \cite{mscnn}. The `FDNN2+SS' has an even better accuracy with an L-AMR of $49.8\%$ under the 'ALL' evaluation setting. The L-AMR vs. FPPI plots for the 'Reasonable' and 'All' evaluation settings are shown in Figure~\ref{fig:cal_res} and Figure~\ref{fig:cal_all}. Except the VJ \cite{Viola} and the HOG \cite{HOG} methods, which are plotted as the baselines, all the other results are CNN-based methods.


\begin{figure}
\begin{center}
   \includegraphics[width=1\linewidth]{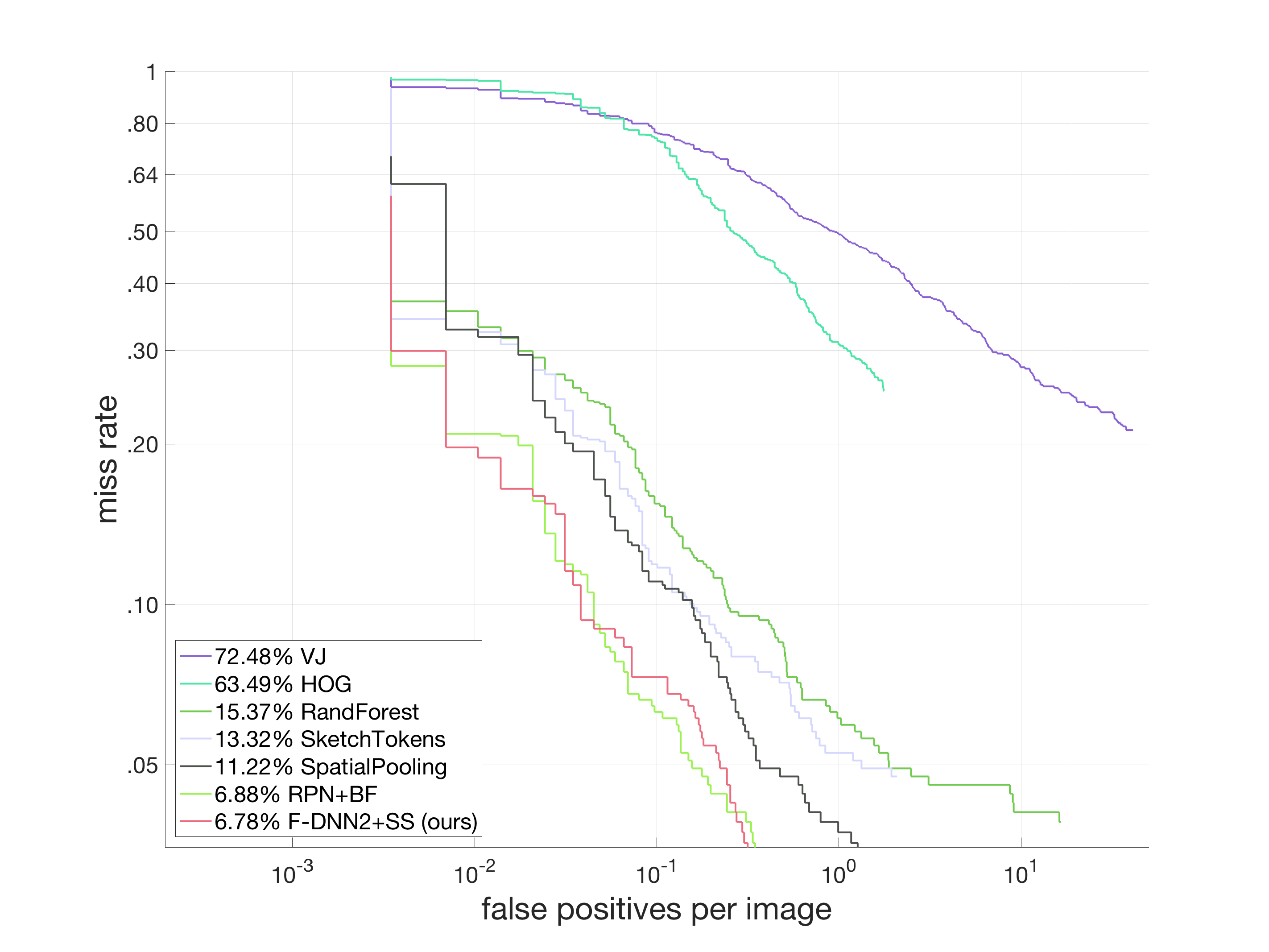}
\end{center}
   \caption{L-AMR vs. FPPI plot on 'Reasonable' evaluation setting on INRIA dataset.}
\label{fig:inria}
\end{figure}

\textbf{Evaluation on INRIA}: We  evaluate the proposed method using the converted INRIA pedestrian dataset provided by Caltech Pedestrian group. There are $614$ full positive training images and $288$ full positive testing images in the INRIA dataset. At least one pedestrian is annotated in each image. To test the generalization capability of our model, we directly test our Caltech-pretrained model on the INRIA test set without any fine-tuning on the INRIA training set. On the 'Reasonable' setting, our `FDNN2+SS' method achieves $6.78\%$ L-AMR, outperforming the previous best result $6.9\%$ by RPN+BF. Table~\ref{tab:inria} shows the best results reported on INRIA dataset and Figure~\ref{fig:inria} shows the L-AMR vs. FPPI plot. Results from VJ and HOG are plotted as the baselines.

\begin{figure}
\begin{center}
   \includegraphics[width=1\linewidth]{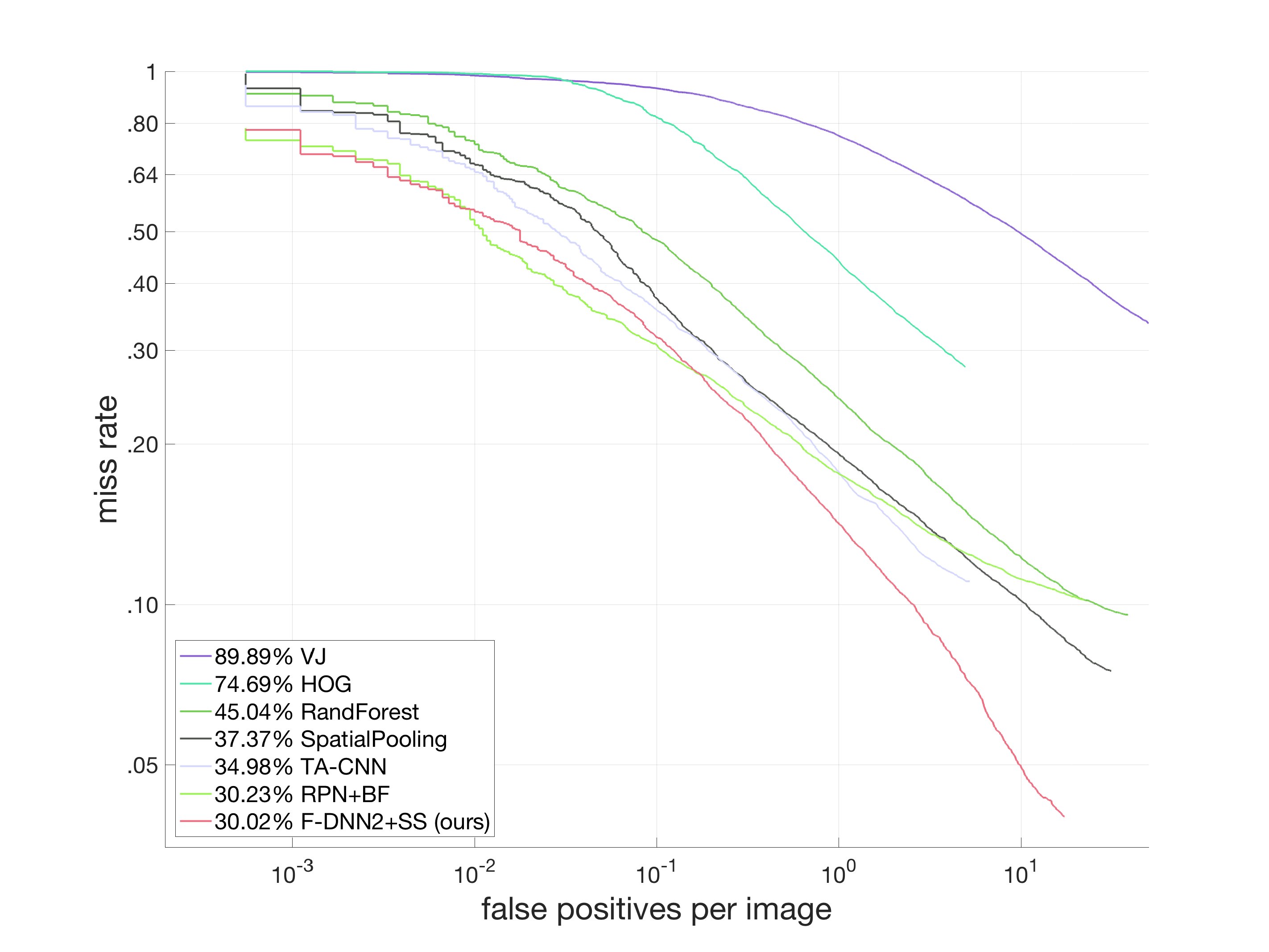}
\end{center}
   \caption{L-AMR vs. FPPI plot on 'Reasonable' evaluation setting on ETH dataset.}
\label{fig:eth}
\end{figure}

\textbf{Evaluation on ETH}: There are $1,804$ images from three video sequences in the ETH pedestrian dataset. As we used the ETH images to train our model, in order to test on the ETH dataset, we removed all the training images from the ETH dataset in our training set and retrained our model. On the 'Reasonable' setting, our method achieves $30.02\%$ L-AMR, outperforming the previous best result $30.23\%$ by RPN+BF. Table~\ref{tab:eth} shows the best results reported on the ETH dataset and Figure~\ref{fig:eth} shows the L-AMR vs. FPPI plot. Results from VJ and HOG are plotted as the baselines.

\textbf{Evaluation on KITTI}: We further generalize our method to multi-class detection problem and test on KITTI object detection dataset. KITTI object detection dataset contains $7,481$ training images and $7,518$ test images. All the annotations are split into 7 classes such as cars, vans, trucks, pedestrians, cyclists, trams, and 'Don't care'. Only cars, pedestrians, and cyclists are evaluated. There are three evaluation settings as shown in Table~\ref{tab:kitti_settings}. Following \cite{mscnn}, we split the training data into a training set and a validation set. We fine-tune three models using all training annotations for the three evaluating classes respectively. For the three models, we set the main aspect ratio to the mean aspect ratio of each class, which is $0.4$ for pedestrians, $0.7$ for cyclists, and $1.6$ for cars. Table~\ref{tab:kitti} shows the results on KITTI object detection dataset. We achieved comparable results on all classes. Since the Caltech Pedestrian dataset doesn't distinguish between pedestrians and cyclists, while the KITTI object detection dataset does, it degrades our performance on the pedestrians class and the cyclists class on KITTI.

\begin{table}[h!]
\begin{center}
\begin{tabular}{|l|l|l|l|l|}
\hline
Benchmark&Easy&Moderate&Hard\\
\hline\hline
Car& 89.68 \% & 85.11 \% & 70.35 \%\\
Pedestrian& 74.05 \% & 61.17 \% & 57.15 \%\\
Cyclist& 67.06 \% & 51.85 \% & 46.67 \%\\
\hline
\end{tabular}
\end{center}
\caption{Evaluation results on KITTI object detection dataset.}
\label{tab:kitti}
\end{table}

\begin{table}[h!]
\begin{center}
\begin{tabular}{|l|l|}
\hline
Method & Reasonable\\
\hline\hline
CG & 13.06\%\\
CG+GglNet & 8.64\%\\
CG+ResNet & 8.38\%\\
CG+GglNet+ResNet+Fusion net (F-DNN2) & 8.12\%\\
CG+GglNet+ResNet+Fusion net+SS (F-DNN2+SS)& 7.67\%\\
\hline
\end{tabular}
\end{center}
\caption{Ablation study of our system.}
\label{tab:netfusion}
\end{table}

\begin{table}[h!]
\begin{center}
\begin{tabular}{|l|l|l|}
\hline
Method & Hard-label & Soft-label\\
\hline\hline
CG + ResNet & 8.97\% & 8.38\%\\
CG + GglNet & 9.41\% & 8.64\%\\
CG + GglNet + ResNet & 8.65\% & 8.12\%\\
\hline
\end{tabular}
\end{center}
\caption{Effectiveness of the soft-label method compared to the conventional hard-label method on Caltech Pedestrian dataset using the reasonable setting.}
\label{tab:slhl}
\end{table}

\begin{table}
\begin{center}
\begin{tabular}{|l|l|}
\hline
Method & Speed on TITAN X \\
&(seconds per image)\\
\hline\hline
CompACT-Deep & 0.5\\
SAF R-CNN & 0.59\\
F-DNN2 & \textbf{0.3}\\
F-DNN2 (Reasonable) & \textbf{0.16} \\
CG+GglNet (Reasonable) & \textbf{0.11} \\
CG+SqueezeNet (Reasonable) & \textbf{0.09} \\
F-DNN2+SS & 2.48\\
\hline
\end{tabular}
\end{center}
\caption{A comparison of speed among the state-of-the-art models.}
\label{tab:speed}
\end{table}

\begin{table*}[h!]
\begin{center}
\begin{tabular}{|l|l|l|l|l|l|l|l|l|}
\hline
Method & Reasonable & All & Far & Medium & Near & Occ. none & Occ. partial & Occ. heavy\\
\hline\hline
CG+GglNet & 8.64\% & 51.59\% & 76.87\% & 37.75\% &  1.72\% & 7.18\% & 18.05\% & 41.19\%\\
\hline
CG+ResNet & 8.38\% & 49.58\% & 74.60\% & 34.88\% &  1.70\% & 6.94\% & 19.26\% & 42.71\%\\
\hline
\end{tabular}
\end{center}
\caption{Breakdown comparisons between CG+GglNet and CG+ResNet on Caltech Pedestrian dataset.}
\label{tab:gglres}
\end{table*}

\subsection{Result analysis}
\subsubsection{An ablation study: effectiveness of the network fusion technique}
In this subsection, we analysis the performance increases step by step from the candidate generator (CG) to the final system. The L-AMR is $13.06\%$ by using the candidate generator alone, due to the large number of false positives. By fusing the candidate generator with GoogLeNet, we can improve the performance to $8.64\%$. By fusing the candidate generator with ResNet-50, we can improve the L-AMR to $8.38\%$. Furthermore, by fusing the candidate generator with GoogLeNet and ResNet-50 using our proposed fusion net, we can achieve the lowest L-AMR so far at $8.12\%$. Finally, by fusing the semantic segmentation network into our system, we can achieve the best performance at $7.67\%$. The analysis shows the capability of the network fusion framework and the advantages of using the idea of ensemble learning. The results of ablation study are given in Table~\ref{tab:netfusion}.

\begin{figure*}
\begin{center}
   \includegraphics[width=1.0\linewidth]{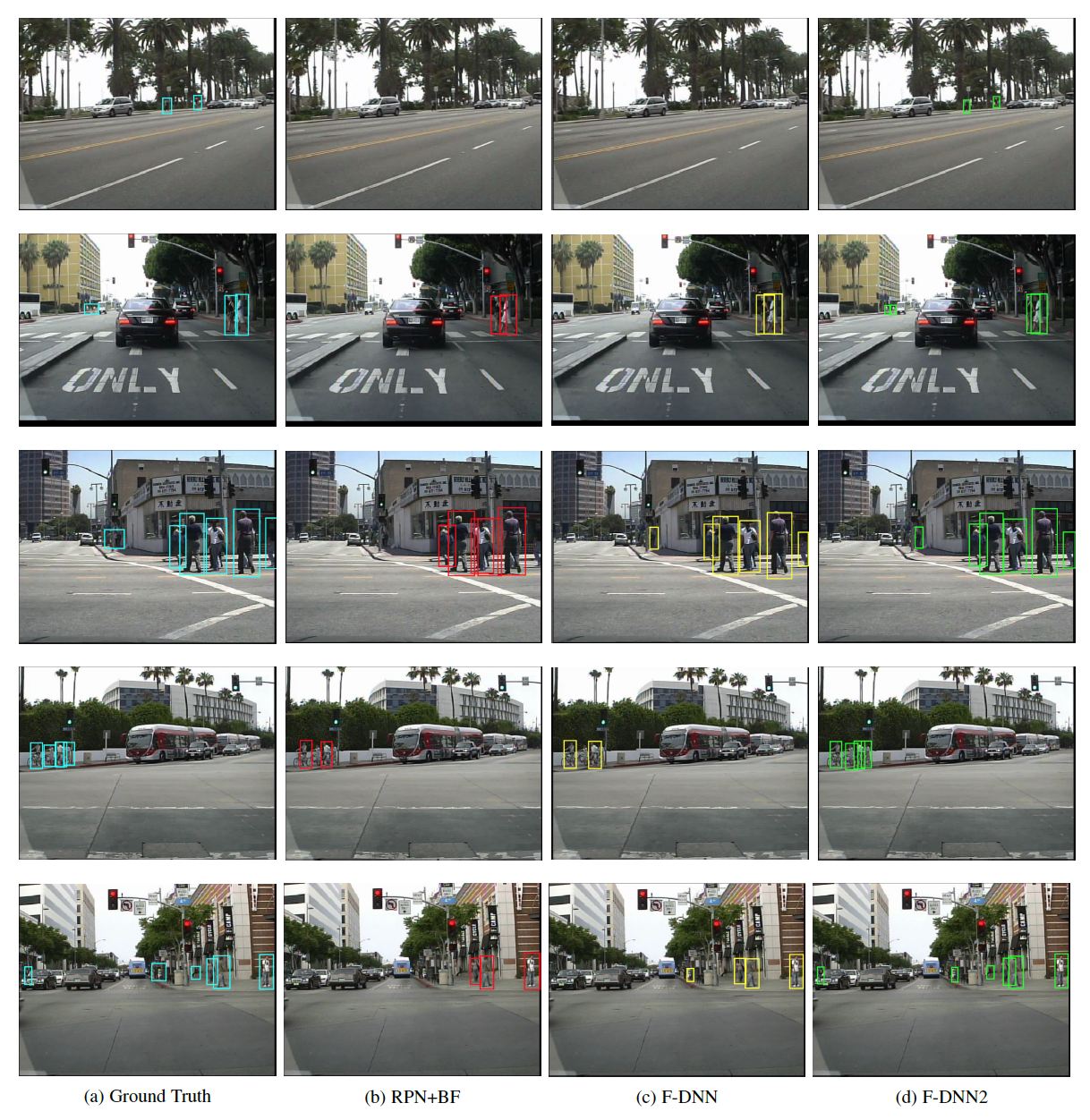}
\end{center}
\caption{Detection comparisons on five challenging pedestrian detection scenarios. Each row represents one challenging scenario and the four columns show the bounding boxes from the ground truth, RPN+BF, F-DNN, and F-DNN2, respectively.}
\label{fig:visualize}
\end{figure*}

\subsubsection{GoogLeNet VS. ResNet-50}
We explore how each part of the classification system contributes to our final results. The breakdown performance comparisons of all evaluation settings on the Caltech Pedestrian dataset between fusing with GoogLeNet alone and fusing with ResNet-50 alone are given in Table~\ref{tab:gglres}. From the results we can see that GoogLeNet works better in partial/heavy occluded pedestrians while ResNet-50 works better in non-occluded pedestrians. By analyzing the weights learnt in Equation~\eqref{eq:10}, we see that the weight for GoogLeNet is $1.11$ and the weight for ResNet-50 is $2.22$, which means that our fusion network values the ResNet-50 more than the GoogLeNet. This is reasonable since there are more non-occluded pedestrians in the training data.

\subsubsection{Soft-label method versus hard-label method}
To test how effectively the soft-label method improves the performance, we compare with the conventional hard-label method on the Caltech Pedestrian dataset. Since we use the overlap ratio between the candidate bounding box and the ground-truth annotation to assign labels, the soft-label method gives us not only the information of the existence of a pedestrian in each candidate's bounding box, but also how much of the bounding box belongs to the pedestrian. This feature benefits even more in cases where the overlap ratio is around $0.5$: e.g. it is too risky to directly assign a hard-label $1$ or $0$ to a bounding box with overlap ratio $0.49$ or $0.51$. We give the performance comparisons between the hard-label method and the soft-label method in Table~\ref{tab:slhl}.

\subsubsection{Results visualization on challenging scenarios and failure cases}
We visualize the detection results generated by our system compared with previous state-of-the-art systems on several challenging scenarios: far pedestrians, crowed scene, occluded pedestrians, pedestrians overlap with each other. Figure~\ref{fig:visualize} visualizes detection results on five challenging scenarios. In Figure~\ref{fig:visualize}, each row represents one challenging scenario and the four columns show the bounding boxes from the ground truth, RPN+BF, F-DNN, and F-DNN2, respectively. From the visualizations we can see that our model is more robust and accurate on various challenging cases.


\subsubsection{Speed analysis}
We use one NVIDIA TITAN X GPU to analysis the processing speed of each component and the overall architecture of our network. There are four main components: the candidate generator, GoogLeNet, ResNet-50, and the semantic segmentation network. Since the candidate generator has a fully convolutional framework which removes the fully connected layers in the original VGG net, it takes $0.06$s to process one image. To test the processing time of the classification system, we have two settings: the first test runs on all pedestrian candidates; The second test runs only on candidates above 40-pixel in height, which is designed for the 'reasonable' evaluation setting. Since the number of pedestrian candidates varies from image to image, the test reports the mean processing time of all images. Running the candidate generator followed by GoogLeNet and ResNet-50 in parallel on one GPU, the speed for the classification system is $0.3$s and $0.16$s per image for the two tests. To achieve real time pedestrian detection, we fuse the candidate generator with SqueezeNet~\cite{SqueezeNet}. The processing time per image is $0.09$s, while being $10.8\%$ in L-AMR. By processing the semantic segmentation network in parallel with the pedestrian detection system and fusing them together, the processing time per image of our final model is $2.48$s. The speed comparisons of our models and other methods are given in Table~\ref{tab:speed}.


\section{Conclusion and Discussion}
We present an effective solution to the pedestrian detection problem in this paper. The proposed system consists of two parallel subsystems: The main pedestrian detection system to generate all detections and the semantic segmentation system to help refine the results. The pedestrian detection system further consists of a pedestrian candidate generator and a classification system. To give more bounding box information to the classification network, we propose a new soft-label method which assigns floating point label to all classes. We implement the idea of ensemble learning to design the classification system. We proposed a soft-rejection network fusion methodology to combine the opinions of all classification networks and the semantic segmentation network with that of the  candidate generator network.

The performance of our system is evaluated on four popular pedestrian detection datasets: Caltech Pedestrian dataset, INRIA dataset, ETH dataset, and KITTI dataset. We achieve the state-of-the-art performance on the first three datasets and comparable results on the KITTI dataset. Our system also works faster in processing speed than other state-of-the-arts when testing using the same experiment settings. The results and analysis show that our system works accurately, efficiently, and robustly to detect pedestrians and other object classes under various challenging scenarios. 

\bibliographystyle{ieeetran}
\bibliography{fdnnjournal2}

\end{document}